\documentclass{article}
\PassOptionsToPackage{numbers, compress}{natbib}
\usepackage[T1]{fontenc}
\usepackage{arxiv}

\usepackage{natbib}
\usepackage{microtype}
\usepackage{graphicx}
\usepackage{subcaption}
\usepackage{booktabs} 
\usepackage{booktabs}
\usepackage{tabularx}
\usepackage{array}
\usepackage{caption}
\newcolumntype{Y}{>{\raggedright\arraybackslash}X}
\usepackage{hyperref}
\usepackage{xcolor}
\usepackage{amsmath}
\usepackage{amssymb}
\usepackage{mathtools}
\usepackage{amsthm}
\usepackage{bm}
\usepackage[capitalize,noabbrev]{cleveref}

\theoremstyle{plain}

\theoremstyle{definition}

\theoremstyle{remark}

\newcommand{\methodname}{\textsc{MosaicMRI}}

\newcommand{\PA}[1]{{#1}}

\DeclareMathOperator*{\argmin}{arg\,min}
\newcommand{\twonorm}[1]{\left\| #1\right\|}
\newcommand{\rbr}[1]{\left(#1\right)}
\newcommand{\bbr}[1]{\left[#1\right]}

\title{\methodname{}: A Diverse Dataset and Benchmark for \\ Raw Musculoskeletal MRI}

\author{%
Paula Arguello$^{1,3}$,
Berk Tinaz$^{2,3}$,
Mohammad Shahab Sepehri$^{2,3}$,\\
\textbf{Maryam Soltanolkotabi}$^{4}$,
\textbf{Mahdi Soltanolkotabi}$^{1,2,3}$\\
$^{1}$Department of Computer Science, University of Southern California\\
$^{2}$Department of Electrical and Computer Engineering, University of Southern California\\
$^{3}$USC Center on AI Foundations for the Sciences\\
$^{4}$Department of Radiology and Imaging Sciences, University of Utah\\
\texttt{parguell@usc.edu}
}

\begin{document}
\maketitle

\begin{abstract}
Deep learning underpins a wide range of applications in MRI, including reconstruction, artifact removal, and segmentation. However, progress has been driven largely by public datasets focused on brain and knee imaging, shaping how models are trained and evaluated. As a result, careful studies of the reliability of these models across diverse anatomical settings remain limited. In this work, we introduce \methodname{}, a large and diverse collection of fully sampled raw musculoskeletal (MSK) MR measurements designed for training and evaluating machine-learning--based methods. \methodname{} is the largest open-source raw MSK MRI dataset to date, comprising 3,150 volumes and 89,830 slices, including a 3,096-volume 1.5~T core dataset and a 54-volume 0.55~T low-field set. The dataset offers substantial diversity in volume orientation (e.g., axial, sagittal), imaging contrasts (e.g., PD, T1, T2), anatomies (e.g., spine, knee, hip, ankle, and others), numbers of acquisition coils, and field strengths. Using accelerated reconstruction as a testbed, we study scaling, robustness, and data selection under realistic MSK distribution shifts. Across E2E-VarNet, U-Net, and ViT baselines, mixed-anatomy training improves over anatomy-specific training, showing that the benefit of anatomical diversity is not architecture-specific. Zero-shot experiments on the 0.55~T low-field set highlight that using more training data improves average reconstruction under field-strength shift. Controlled subset experiments further suggest that these gains come from transferable cross-anatomy structure, rather than anatomy or contrast matching alone. \footnote{\methodname{} includes raw measurements and acquisition metadata. The data were collected with appropriate patient consent under IRB approval, including authorization for public release. The \methodname{} dataset and accompanying benchmark are available at: \url{https://mosaicmri.ai}. Codebase is available at: \url{https://github.com/AIF4S/mosaicmri}.}
\end{abstract}

\section{Introduction}  \label{sec:intro}

Magnetic resonance imaging (MRI) is a cornerstone modality in clinical imaging, particularly valued for its superior soft-tissue contrast and multiparametric tissue characterization without exposure to ionizing radiation. It plays a central role in the evaluation of neurologic, musculoskeletal (MSK), and oncologic disease, where subtle differences in tissue composition and microstructure are diagnostically critical. Unlike projection-based imaging modalities, MRI data are acquired in the spatial frequency domain (k-space). Image contrast is not fixed but arises from sequence design and tissue-specific relaxation properties (e.g., T1, T2, proton density), enabling flexible, task-specific contrast mechanisms. This combination of physics-derived encoding and contrast programmability makes MRI uniquely information-rich, but also computationally complex with longer scan times. 

Recent years have seen rapid progress in accelerating MRI through advances in computational methods, particularly those based on compressed sensing \citep{candes_stable_2006, donoho_compressed_2006} and, more recently, deep learning \citep{hammernik2018learning, sriram_end--end_2020, fabian2022humus}. By exploiting structure and redundancy in MR data, learning-based reconstruction methods have demonstrated impressive improvements in image quality at high acceleration factors, significantly outperforming classical approaches. These methods are now widely studied for tasks such as accelerated reconstruction, artifact suppression, segmentation, and other downstream analyses \citep{ronneberger2015u, dalmaz2022resvit, desai2022skmteadatasetacceleratedmri, fabian2022humus}.

Despite this progress, the development and evaluation of deep learning methods for MRI have been heavily shaped by the availability of public datasets. Much like the ``ImageNet moment" in computer vision \citep{russakovsky2015imagenetlargescalevisual}, large-scale, openly available raw MRI datasets have been instrumental in driving methodological innovation \cite{zbontar_fastmri_2019}. However, most of these focus on a narrow set of body parts (with heavy emphasis on the brain and knee, see Table \ref{tab:dataset_comparison}). This focus biases model design, training, and evaluation towards a limited anatomical scope implicitly. As a result, relatively little is known about how well current learning-based MRI methods scale, generalize, and remain robust when applied to more diverse anatomical settings encountered in clinical practice, such as MSK MRI. MSK MRI spans a diverse set of joints and anatomic regions, including the spine and peripheral joints such as the shoulder, hip, knee, ankle, and wrist, each with distinct biomechanics, tissue composition, and clinical indications for imaging. Imaging protocols vary substantially across each of these anatomical regions, with differences in fields of view, spatial resolution, coil selection, and motion susceptibility, depending on whether the goal is to evaluate cartilage, marrow, tendons, ligaments, or postoperative changes. This heterogeneity in anatomy, tissue contrast requirements, and acquisition strategy makes MSK MRI technically demanding while remaining central to the diagnosis of degenerative, traumatic, and neoplastic conditions. Despite its clinical importance, large-scale access to raw k-space MSK datasets remains limited. Compared to neuroimaging, MSK MRI is more fragmented across joints, vendors, and protocol designs, which has slowed the development and \PA{systematic} evaluation of learning-based methods in this domain.

\begin{figure}[!t]
    \centering
    \begin{minipage}{0.43\linewidth}
        \centering
        \includegraphics[width=\linewidth]{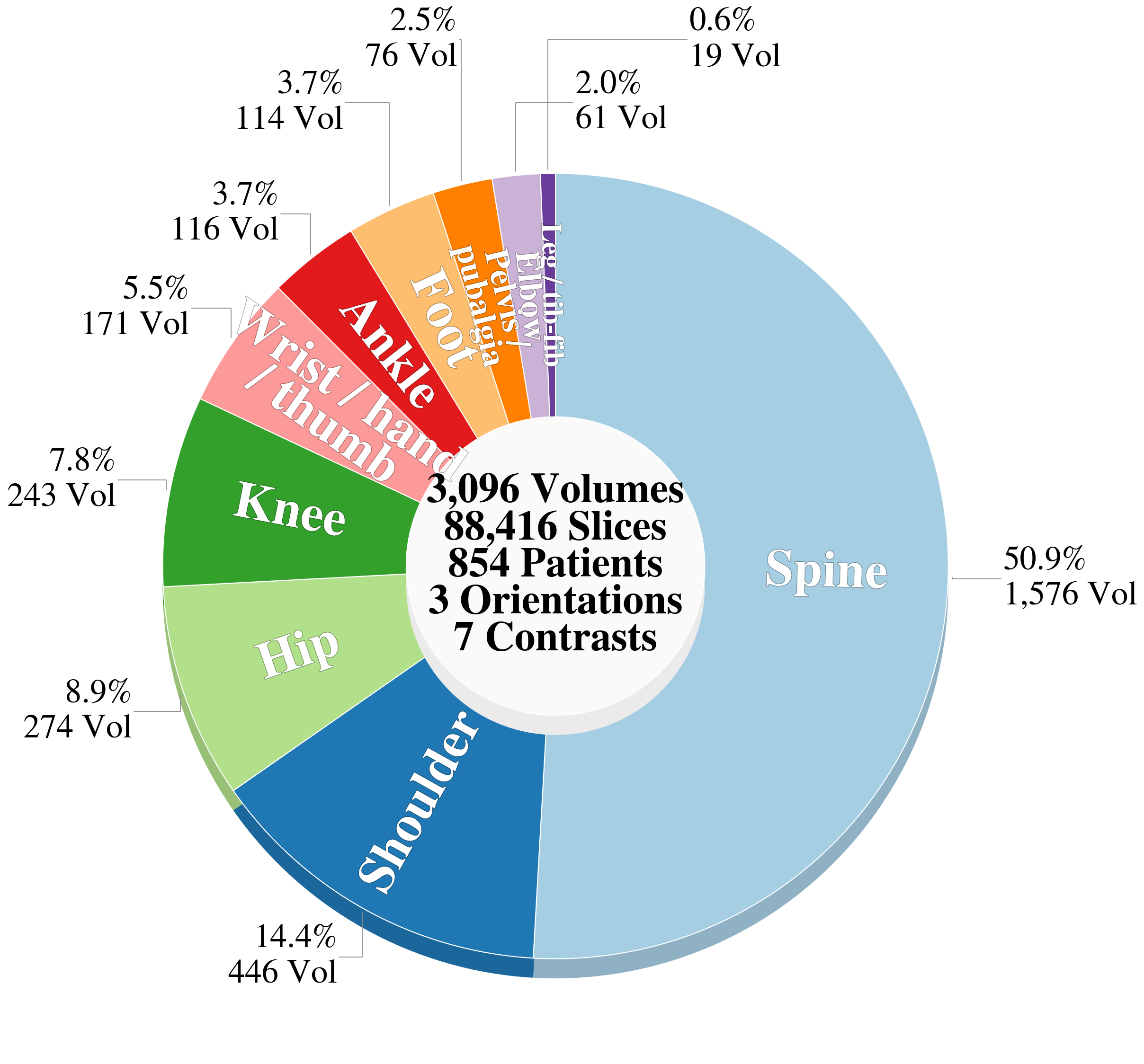}
    \end{minipage}
    \hfill
    \begin{minipage}{0.56\linewidth}
        \centering
        \includegraphics[width=\linewidth]{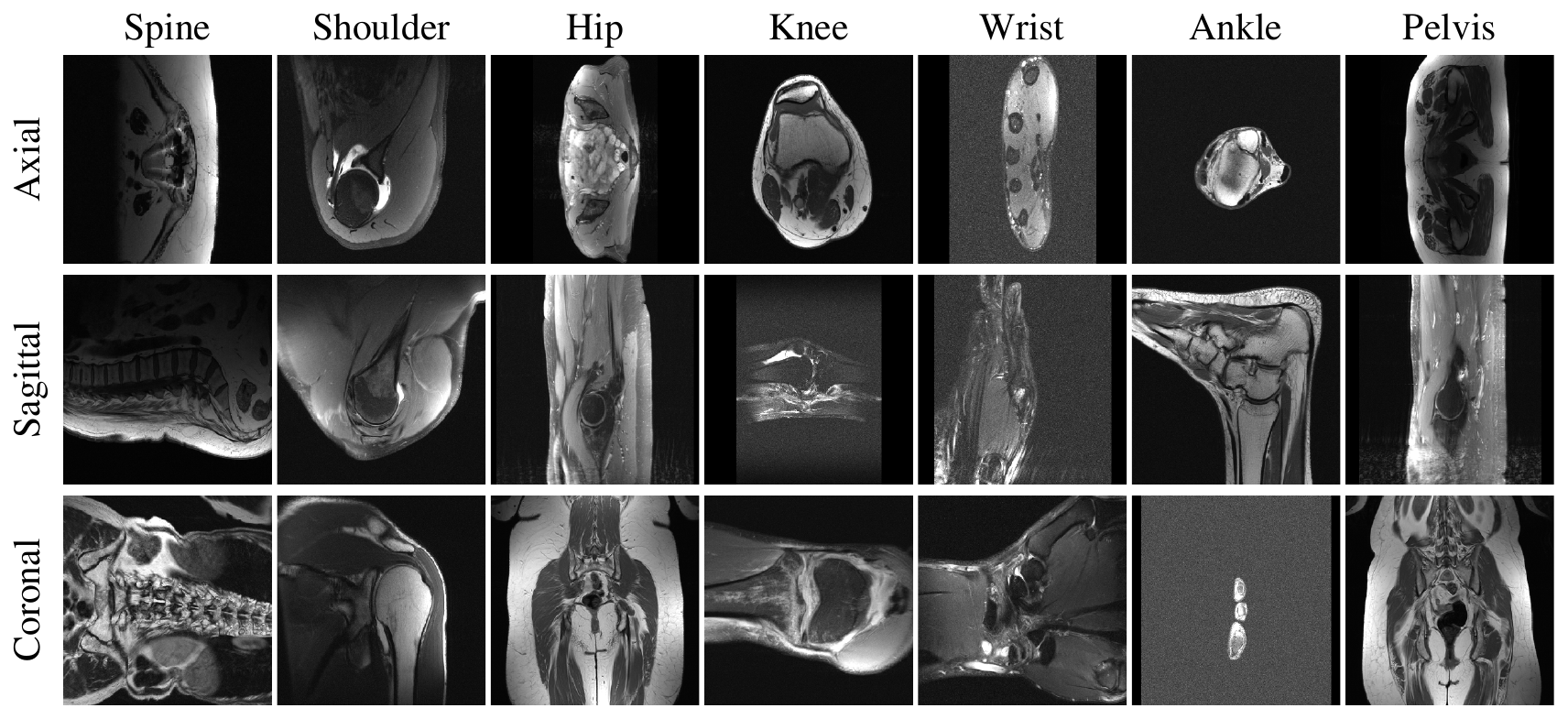}
    \end{minipage}
    
    \caption{\textbf{MosaicMRI overview.}
    \textbf{(left)} Anatomy distribution for 3,096 fully sampled 1.5~T volumes, led by spine (50.9\%, 1,576 volumes), shoulder (14.4\%, 446), and hip (8.9\%, 274), with all ten anatomy categories shown separately.
    \textbf{(right)} Representative slices spanning seven anatomy groups and three orientations.}
    
    \label{fig:mosaic_overview}
\end{figure}

To address these gaps, we introduce \methodname{}, a large-scale, open-source collection of fully sampled raw musculoskeletal MRI data designed to support systematic research in learning-based MRI methods. \methodname{} substantially expands the range of clinically relevant MSK examinations represented in existing public datasets, incorporating multiple joints, protocol-specific contrasts, imaging planes, and coil configurations encountered in routine practice. Using \methodname{}, we conduct an extensive experimental study centered on accelerated MRI reconstruction. Employing E2E-VarNet \citep{sriram_end--end_2020} as a representative baseline, we analyze reconstruction performance as a function of both dataset size and model capacity. We further evaluate cross-anatomy generalization by training models on one anatomy and testing them on others, providing new insights into the structure of anatomical similarity and transferability within the MSK domain.

Our main contributions can be summarized as follows:
\begin{itemize}

    \item \PA{We introduce \methodname{}, the largest open-source raw musculoskeletal MRI dataset to date, comprising $3{,}096$ fully sampled 1.5~T multi-coil volumes ($88{,}416$ slices) from 854 patients, plus 54 low-field volumes ($1{,}414$ slices) from 8 patients. The data span a wide range of MSK sites, contrast weightings, imaging planes, and coil configurations. In terms of volume count, \methodname{} is more than $2.5\times$ the size of the largest previously available public MSK MRI dataset and moves beyond brain- and knee-centric benchmarks.}

    \item \PA{We present a comprehensive empirical study of accelerated MRI reconstruction under realistic MSK variability. Across E2E-VarNet, U-Net, and ViT baselines, larger and more diverse training sets consistently improve reconstruction quality, with especially clear gains in low-data regimes and in zero-shot evaluation on the $0.55$~T low-field set.}

    \item \PA{We analyze cross-anatomic generalization and robustness, identifying groups of body parts (e.g., \texttt{foot} and \texttt{elbow}) that transfer well to each other. We further show that performance under domain shift depends jointly on training-set size, anatomy, and protocol-specific factors.}

    \item \PA{Inspired by recent work \citep{lin2025improvingdeeplearningaccelerated}, we show that similarity-based filtering in DreamSim embedding space can identify useful cross-anatomy supervision under matched training-slice budgets.}

\end{itemize}

\section{Related Work} \label{sec:related_work}

\textbf{Public MRI datasets} -- The fastMRI dataset \citep{zbontar_fastmri_2019} is one of the largest publicly available raw MRI datasets, focusing on two anatomies (knee and brain). It contains 1.2k multi-coil knee volumes and 6.4k brain volumes and has become a de facto benchmark with public leaderboards. Its establishment spurred community-wide interest and challenge competitions in accelerated MRI reconstruction \citep{muckley2021results}. Recently, the fastMRI initiative was extended to breast MRI for dynamic contrast-enhanced (DCE) scans (fastMRI breast \citep{solomon2025fastmri}) and to prostate MRI (fastMRI prostate \citep{tibrewala2024fastmri}) for cancer imaging.

Beyond fastMRI, several other open datasets are commonly used. Early on, NYU knee dataset \citep{hammernik2018learning} and Stanford's fully-sampled MRI datasets (2D FSE, and 3D FSE knee) \citep{website:Stanford2D} provided testbeds for learning-based reconstruction despite their small size (with 20 to 100 volumes). More recently, OCMR \citep{chen2020ocmr} and CMRxRecon2023 \citep{wang2024cmrxrecon} datasets consist of multi-coil cardiac MR data,  and SKM-TEA \citep{desai2022skmteadatasetacceleratedmri} multi-coil knee scans with manual segmentation masks and bounding box annotations for clinically relevant pathologies. AHEAD \citep{caan2022ahead} provided motion-corrected 7T brain MRI scans across the adult lifespan, extending public data to ultra-high-field imaging. Meanwhile, the M4Raw dataset \citep{lyu2023m4raw} tackles the opposite end of the field-strength spectrum: it contains multi-contrast brain MRI scans acquired at 0.3T to facilitate research in low-field MRI reconstruction.

Despite this growing list of datasets, comprehensive MSK MRI data remain largely absent. Before our work and to the best of our knowledge, the only multi-anatomy MSK collections are the SMURF \citep{bachrata_smurf} (which includes knee, breast, and abdomen) and Stanford 2D FSE \citep{website:Stanford2D} (includes lower extremities and pelvis). However, SMURF and Stanford 2D FSE datasets contain only 113 and 89 volumes, respectively, making them relatively small for training and evaluation. In short, existing public MRI datasets have been either large but anatomically narrow or anatomically diverse but small. \methodname{} addresses this need by providing the largest open collection of raw MSK MRI to date. In Table \ref{tab:dataset_comparison}, we contrast the coverage of aforementioned datasets with \methodname{}. Notably, \methodname{} is more than 2.5 times the size of the largest comparable dataset (fastMRI knee) with substantially more diversity in clinically relevant MSK anatomies.

\begin{table*}[t]
\centering
\caption{Comparison of public MRI datasets with raw k-space data. Statistics are compiled from respective dataset papers and Table~1 of \citep{lin2025improvingdeeplearningaccelerated}.}
\label{tab:dataset_comparison}

\scriptsize
\begin{tabularx}{\textwidth}{lYcYccccc}
\toprule
Dataset
& Anatomy
& View
& Contrast
& Vendor
& Magnet
& Coils
& \# Scans / \# Subjects \\
\midrule
fastMRI brain \citep{zbontar_fastmri_2019}
& brain
& axial
& T1, T1POST, T2, FLAIR
& Siemens
& 1.5T, 3T
& 4-20
& 6.4k / 6.4k \\

OCMR \citep{chen2020ocmr}
& heart
& various
& SSFP
& Siemens
& 0.5T - 3T
& 15-38
& 165 / 165 \\

AHEAD \citep{caan2022ahead}
& brain
& various
& MP2RAGE-ME
& Philips
& 7T
& 32
& 105 / 105  \\

M4Raw \citep{lyu2023m4raw}
& brain
& axial
& T1, T2, FLAIR
& XGY
& 0.3T
& 4
& 1.3k / 183 \\

CMRxRecon2023 \citep{wang2024cmrxrecon}
& heart
& various
& SSFP-Balanced
& Siemens
& 3T
& 10
& 300 / 300 \\

fastMRI prostate \citep{tibrewala2024fastmri}
& prostate
& axial
& T2, DWI
& Siemens
& 3T
& 10-30
& 312 / 312 \\

fastMRI breast \citep{solomon2025fastmri}
& breast
& various
& VIBE
& Siemens
& 3T
& 16
& 300 / 284 \\

\midrule

Stanford 2D FSE \citep{website:Stanford2D}
& lower extremity, pelvis, and others
& various
& PD 
& GE
& 3T
& 3-32
& 89 / 89 \\

NYU dataset \citep{hammernik2018learning}
& knee
& various
& PD, PDFS, T2FS
& Siemens
& 3T
& 15
& 100 / 20 \\

fastMRI knee \citep{zbontar_fastmri_2019}
& knee
& coronal
& PD, PDFS
& Siemens
& 1.5T, 3T
& 15
& 1.2k / 1.2k \\

SMURF \citep{bachrata_smurf}
& knee, breast, abdomen
& various
& FSE, FatSat, WatSat, Dixon
& Siemens
& 3T
& 10-20
& 113 / 11 \\

SKM-TEA \citep{desai2022skmteadatasetacceleratedmri}
& knee
& various
& qDESS
& GE
& 3T
& 8, 16
& 155 / 155 \\

\midrule
\textbf{MosaicMRI (ours)}
& all MSK anatomies
& various
& T1, T1FS, T2, T2FS, PD, PDFS, STIR
& Siemens
& \PA{1.5T; 0.55T}
& \PA{4-46}
& \PA{3.15k / 862} \\

\bottomrule
\end{tabularx}

\end{table*}

\textbf{Learning-based methods in MRI} -- Machine learning now underpins a broad array of applications in MRI. On the image reconstruction front, learning-based methods have advanced accelerated MRI \PA{and are also applied} to tasks like artifact correction. For instance, networks have been developed to suppress motion artifacts or Gibbs ringing in MR images, outperforming traditional post-processing. Deep learning has also been used for image enhancement tasks such as super-resolution (e.g., reconstructing high-resolution images from lower resolution scans) and for contrast synthesis \citep{chartsias_adversarial_2017, dar2019mrisynth, dalmaz2022resvit}, where one MRI contrast (such as T2-weighted) is synthesized from another (such as T1-weighted) using learned models. Such applications can assist in improving scan times and increasing the diversity of diagnostic information when certain sequences are missing or of poor quality.

Beyond improving images themselves, deep learning drives many downstream analysis tasks on MRI. A prime example is the segmentation of anatomical structures or pathologies. Since the introduction of the U-Net (a convolutional network architecture that excelled in biomedical image segmentation), CNN-based models have become the standard for delineating tissues like brain tumors, cardiac structures, or knee cartilage on MR images \citep{ronneberger2015u, desai2022skmteadatasetacceleratedmri}.

\textbf{Robustness and generalization in MRI} -- Multiple studies have demonstrated that deep learning models for accelerated MRI are
vulnerable to distribution shifts. For example, \citet{johnson2021fastmrirobustness} reported that the models submitted to the fastMRI challenge \citep{knoll_assessment_2019} exhibit degraded performance when evaluated on data from different distributions. Similarly, \citet{darestani21arobustness} observed that reconstruction methods for MRI, whether data-driven or hand-tuned, show comparable drops in performance when confronted with distribution shifts. More recently, \citet{lin2024robustnessdeeplearningaccelerated} systematically explored how the composition of training data influences both in-distribution and out-of-distribution reconstruction performance. Specifically, they show that training reconstruction models on datasets that combine images from different scanners, anatomies, contrasts, and field strengths leads to robustness that matches or exceeds that of models trained on any single distribution. However, due to the limited availability of large-scale public datasets, their (as well as that of prior work) empirical evaluation is predominantly confined to brain and knee MRI, leaving the robustness of learned reconstruction models for other MSK anatomies underexplored.

\section{\PA{Background: Deep learning for Accelerated MRI Reconstruction}} \label{apx:dl_recon_baselines}

\PA{We give a primer on accelerated MRI reconstruction and classical solvers in Appendix~\ref{apx:acc_mri_primer}. 
Data-driven approaches for reconstruction often replace the hand-designed regularizer with a learned image prior. U-Net-style encoder--decoder networks \citep{ronneberger2015u} provide a convolutional image-domain baseline, using local filtering and multi-scale skip connections to recover spatial detail. E2E-VarNet \citep{sriram_end--end_2020} combines this type of learned convolutional prior with an unrolled reconstruction procedure: each cascade applies a learned regularization update together with an explicit data-consistency step in k-space. At cascade $t$, we write}

\begin{equation}
\label{eq:kspace_iters}
\PA{
\bm{\hat{k}}^{t+1}
=
\bm{\hat{k}}^{t}
-
\mu^t \bm{M}\rbr{\bm{\hat{k}}^{t}-\bm{\tilde{k}}}
+
\mathcal{G}\rbr{\bm{\hat{k}}^{t}},
}
\end{equation}
\PA{where $\bm{\hat{k}}^{t}$ is the current k-space estimate, $\bm{\tilde{k}}$ is the acquired undersampled k-space, $\bm{M}$ is the sampling mask, $\mu^t$ is a learned step size, and $\mathcal{G}(\cdot)$ denotes the learned regularization update.} \PA{We also consider two image-domain alternatives that are not built around explicit k-space data consistency. Vision Transformers (ViTs) \citep{dosovitskiy2021image} represent an image as a sequence of patch tokens and use self-attention to model long-range spatial dependencies. Latent Bridge Matching (LBM) \citep{chadebec2025lbmlatentbridgematching} is a recent latent-space image-to-image translation model designed for one-step or few-step generation.}

\section{The \methodname{} Dataset} \label{sec:dataset}

In this section, we introduce our musculoskeletal (MSK) MRI dataset designed for training and evaluation of learning-based methods under realistic clinical variability. In contrast to existing public raw MRI datasets (see Table \ref{tab:dataset_comparison} for a detailed comparison), \methodname{} exhibits substantially greater heterogeneity in anatomies, contrast weighting, imaging plane, and coil configuration.

\subsection{Constructing the Dataset} \label{sec:dataset_sources}

The source data consists of approximately $4$~TB of anonymized clinical acquisitions collected on a $1.5$~T Siemens Magnetom Avantofit scanner between July~15,~2025 and September~23,~2025. \PA{We additionally include a $0.55$~T low-field MSK set containing 54 scans and 1,414 slices, acquired with a Siemens MAGNETOM Free.Max around April~17,~2026.} From the raw data, we excluded incomplete or aborted scans, non-diagnostic localizer and planning acquisitions, and system calibration scans (e.g., noise reference and coil sensitivity calibrations). Specifically, we exclude protocols incompatible with standard slice-based reconstruction, including calibration-only scans, large 3D sequences (such as SPACE and VISTA), SEMAC, and other volumetric acquisitions. We visually perform quality checks on the remaining scans to remove cases with severe motion and susceptibility artifacts that would hinder downstream analysis. All retained raw k-space data are stored in Hierarchical Data Format Version 5 (HDF5) files, with acquisition metadata encoded in an ISMRMRD-compatible \citep{inati2017ismrm} header structure. The internal layout of the k-space data follows the fastMRI convention, enabling reuse of existing codes for fastMRI on \methodname{} with minimal modifications. 

The curated dataset contains routine slice-based MSK sequences spanning common contrast mechanisms and fat-suppression strategies. The seven contrast groups are T1, T1-FS, T2, T2-FS, PD, PD-FS, and STIR. Each public scan is annotated with orientation, contrast, a ten-class anatomy label, and complementary coarse and fine anatomy labels. \PA{The filtered corpus contains $3{,}096$ fully sampled $1.5$~T volumes ($88{,}416$ slices) from 854 patients, plus the 54-volume $0.55$~T set ($1{,}414$ slices) from 8 patients, for $3{,}150$ volumes and $89{,}830$ slices in total.}

\subsection{Anatomical Coverage} \label{sec:dataset_anatomy}

The dataset encompasses a broad spectrum of MSK MRI examinations, including spine studies (1,576 volumes) as well as peripheral joint imaging of the knee (243), shoulder (446), hip (274), ankle (116), elbow (61), wrist/hand/thumb (171), and foot (114), along with dedicated examinations of the leg/tib-fib (19) and pelvis/pubalgia (76). Figure~\ref{fig:mosaic_overview} shows the volume distribution across anatomies.

\subsection{Acquisition Geometry and Coil Configuration} \label{sec:dataset_geometry}

Acquisition geometry and coil configuration vary widely across the dataset. Rectangular public k-space has 512--1,024 readout samples and 144--679 retained phase-encoding columns. Root sum of squares (RSS) targets span 256--768 pixels in height and 144--560 pixels in width. Volumes contain 12--66 slices (mean 28.56), and receiver-coil counts span 4--46 (mean 15.09); 16 coils are most frequent (1,088 volumes). We share additional statistics in Appendix \ref{apx:dataset_statistics}.

\subsection{Dataset Partitioning} \label{sec:dataset_splits}

\begin{table}[ht]
\begin{minipage}{0.46\linewidth}

To avoid patient-level leakage, Table \ref{tab:dataset_splits} partitions the $1.5$~T core by patient into \texttt{train}, \texttt{val}, and \texttt{test} splits with ratios $70\%,15\%,15\%$. Splits are selected with an objective that jointly balances volumes, slices, patients, anatomies, and contrasts; there is no patient overlap across splits.
\end{minipage}\hfill
\begin{minipage}{0.52\linewidth}

\centering
\footnotesize             
\caption{\PA{Dataset statistics for the $1.5$~T splits and the $0.55$~T out-of-distribution set.}}
\label{tab:dataset_splits}

\resizebox{\linewidth}{!}{
\begin{tabular}{lccc}
\toprule
\textbf{Dataset split} & \textbf{\# scans} & \textbf{\# patients} & \textbf{\# slices} \\
\midrule
\texttt{train}    & 2{,}166 & 598 & 61{,}903 \\
\texttt{val}  & 465     & 128  & 13{,}265 \\
\texttt{test}      & 465     & 128  & 13{,}248 \\
\PA{\texttt{0.55T low-field}} & \PA{54} & \PA{8} & \PA{1{,}414} \\
\bottomrule
\end{tabular}
}
    
\end{minipage}    
\end{table}

\section{Experiments and Results} \label{sec:exp}

In this section, we present results on \methodname{}. We first benchmark reconstruction methods for 4$\times$ and 8$\times$ accelerated multi-coil reconstruction, then \PA{study data scaling, model scaling, cross-anatomy transfer, protocol robustness, dataset and field-strength shift, and similarity-based} filtering to quantify the value of anatomical and protocol diversity.

\subsection{Benchmark Results} \label{sec:exp_benchmark}

We benchmark various methods on \methodname{} using the multi-coil reconstruction task at $4$-fold and $8$-fold acceleration as a testbed. We report standard distortion metrics: peak signal-to-noise ratio (PSNR) and structural similarity index measure (SSIM) on the held-out test split of 465 volumes and 13,248 slices (Table~\ref{tab:mosaic_results_4x_8x}), using the root sum of squares (RSS) reconstruction as the target image. All learned baselines are trained on the \methodname{} training split with random undersampling masks, mirroring benchmarks on FastMRI. Specifically, we randomly sample $25\%$ ($12.5\%$) of whole k-space lines along the phase encoding dimension while retaining $8\%$ ($4\%$) of the lowest frequency band for $4$-fold and $8$-fold acceleration, respectively. Test masks are deterministically seeded by filename. For test set numbers, we evaluate the checkpoint that achieves the highest SSIM on the validation split.

\PA{As a classical baseline, we evaluate ESPIRiT \citep{uecker_espirit_2014} using the \textsc{BART} toolkit \citep{uecker2015berkeley}. As learned baselines, we evaluate E2E-VarNet \citep{sriram_end--end_2020}, U-Net \citep{ronneberger2015u}, ViT \citep{dosovitskiy2021image}, and LBM \citep{chadebec2025lbmlatentbridgematching}. Due to training cost, LBM is evaluated only at $4\times$ acceleration. Additional experimental details are provided in Appendix~\ref{apx:exp_details}.}

\begin{table}[ht]
\caption{Test-set reconstruction of various methods on \methodname{} at $4\times$ and $8\times$ acceleration. Values are mean $\pm$ standard deviation over 465 volumes.}
\label{tab:mosaic_results_4x_8x}
\centering

\setlength{\tabcolsep}{4pt}        
\renewcommand{\arraystretch}{1.1}  
\resizebox{0.7\linewidth}{!}{
\begin{tabular}{@{}lcccc@{}}
\toprule
& \multicolumn{2}{c}{$4\times$ accel.} & \multicolumn{2}{c}{$8\times$ accel.} \\
\cmidrule(lr){2-3} \cmidrule(lr){4-5}
Method 
& PSNR (dB)$\uparrow$ & SSIM$\uparrow$ 
& PSNR (dB)$\uparrow$ & SSIM$\uparrow$ \\
\midrule
ESPIRiT 
& $32.34\!\pm\!5.21$ & $0.627\!\pm\!0.193$ 
& $30.75\!\pm\!4.44$ & $0.598\!\pm\!0.189$ \\
E2E-VarNet 
& $\textbf{41.62}\!\pm\!4.25$ & $\textbf{0.946}\!\pm\!0.046$ 
& $\textbf{38.57}\!\pm\!4.01$ & $\textbf{0.924}\!\pm\!0.054$ \\
\PA{U-Net}
& -- & -- 
& \PA{$37.44\!\pm\!4.34$} & \PA{$0.913\!\pm\!0.056$} \\
\PA{ViT}
& -- & -- 
& \PA{$36.61\!\pm\!4.36$} & \PA{$0.904\!\pm\!0.059$} \\
\midrule
LBM (1 step)
& $33.12\!\pm\!3.13$ & $0.844\!\pm\!0.076$ 
& -- & -- \\
LBM (4 step)
& $32.71\!\pm\!3.08$ & $0.835\!\pm\!0.078$ 
& -- & -- \\
\bottomrule
\end{tabular}
}

\end{table}

We present the benchmark results in Table \ref{tab:mosaic_results_4x_8x}. In line with previous works, we observe that the learning-based baseline E2E-VarNet performs significantly better than the classical baseline ESPIRiT across all metrics and acceleration rates. E2E-VarNet obtains $41.62\pm4.25$~dB PSNR and $0.946\pm0.046$ SSIM at $4\times$, and $38.57\pm4.01$~dB and $0.924\pm0.054$ at $8\times$. We find the LBM numbers to be lower in terms of PSNR ($\approx33$~dB), which is comparable to ESPIRiT, but reasonable in terms of SSIM ($\approx0.84$). We hypothesize that this is due to the perception-distortion tradeoff \citep{Blau_2018}. Although LBM admits excellent perceptual quality, due to the lack of data consistency objective in the training, it produces hallucinated reconstructions that are not consistent with the measurement, which would explain why the PSNR metric is not comparable to that of E2E-VarNet. We provide examples of reconstructed slices in Appendix \ref{apx:vis_recon} (Fig. \ref{fig:recon_visualization_4x_8x}).

\subsection{Scaling Dataset and Model Capacity for MRI Reconstruction} \label{sec:scaling_laws}

\begin{figure*}[!h]
    \centering
    \includegraphics[width=0.95\linewidth]{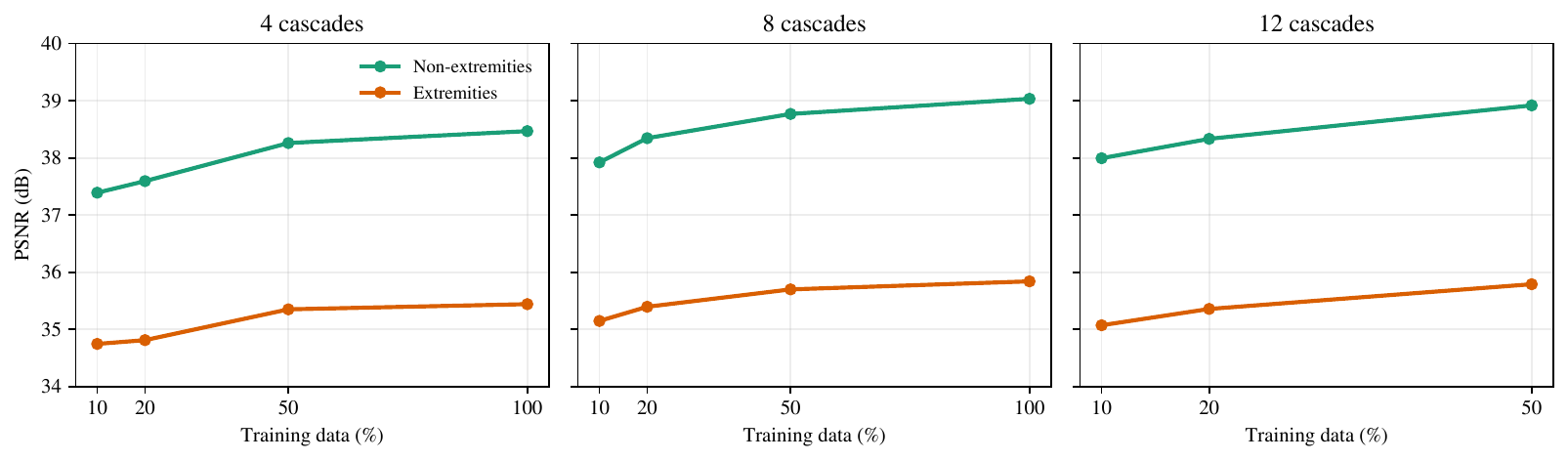}
    
    \caption{PSNR versus training-set fraction for E2E-VarNet with 4, 8, and 12 cascades, stratified into extremity and non-extremity test volumes. All panels use the same y-axis scale.}
    
    \label{fig:scaling_anatomies_plot_casc8_and_12}
\end{figure*}

\methodname{} includes anatomy groups with varying training set sizes. We study how reconstruction quality scales with additional data during training. Given the strong benchmark performance of E2E-VarNet in Section \ref{sec:exp_benchmark}, we train it on varying dataset sizes and model capacity combinations. Specifically, we use fractions $\bbr{0.1,0.2,0.5,1.0}$ of the \texttt{train} and \texttt{val} splits, vary the number of cascades from $\bbr{4,8,12}$, and evaluate on the complete \texttt{test} split with the same deterministic masks.

Figure~\ref{fig:scaling_anatomies_plot_casc8_and_12} shows that PSNR improves as the training fraction increases for the evaluated 4-, 8-, and 12-cascade E2E-VarNet models. This trend is maintained for extremity and non-extremity test volumes, although non-extremity performance remains higher across the evaluated settings. Overall, scaling yields consistent gains, while increasing model capacity further improves reconstruction with diminishing returns.

\begin{table}[ht]
\begin{minipage}{0.5\linewidth}

\PA{Table~\ref{tab:arch_scaling} evaluates U-Net \citep{ronneberger2015u} and ViT \citep{dosovitskiy2021image} under the same training-data fractions. Increasing the training set from $10\%$ to $100\%$ improves PSNR by $1.66$~dB for U-Net and $3.88$~dB for ViT. Both convolutional and transformer baselines improve as the training distribution becomes broader.}

\end{minipage}\hfill
\begin{minipage}{0.48\linewidth}

\centering
\small
\setlength{\tabcolsep}{5pt}
\caption{\PA{Architecture scaling with \methodname{} training data for the held-out test split.}}
\label{tab:arch_scaling}

\resizebox{\linewidth}{!}{
\begin{tabular}{lcccc}
\toprule
Model & 10\% & 20\% & 50\% & 100\% \\
\midrule
U-Net & 35.78 / 0.898 & 36.36 / 0.903 & 36.99 / 0.909 & \textbf{37.44 / 0.913} \\
ViT   & 32.73 / 0.853 & 32.04 / 0.837 & 35.35 / 0.890 & \textbf{36.61 / 0.904} \\
\bottomrule
\end{tabular}
}
    
\end{minipage}    
\end{table}

\subsection{Anatomy Robustness and Generalization} \label{sec:anatomy_generalize}

The scaling trends differ substantially across anatomies, suggesting that data quantity alone does not explain generalization. We therefore evaluate cross-anatomy transfer by comparing E2E-VarNet models trained on a single anatomy to a single model trained on all \textsc{\methodname{}} anatomies (Fig.~\ref{fig:anatomy_robustness}), with rows and columns ordered from higher to lower volume counts. The all-anatomies model achieves the highest overall average (38.57~dB) and is strongest or near strongest across the individual test anatomies. This gain generally persists even when training and test anatomies match, indicating that mixing anatomies helps rather than hurts performance.

To interpret transfer, we highlight (per row) all single-anatomy models that achieve performance within 1~dB of the best single-anatomy result for that target. Blue cells correspond to \emph{high-data anchor} anatomies (\texttt{spine}, \texttt{knee}, \texttt{shoulder}, \texttt{hip}) tested between themselves. Models trained on these anchors achieve the strongest single-anatomy averages and are rarely improved upon by training on a different anchor. However, transfer between anchors is limited, suggesting distinct acquisition regimes despite large sample sizes. These models generalize broadly to many of the smaller anatomies (purple area), indicating that training-set size can be a dominant factor for transfer.

Another important group, comprising \texttt{ankle}, \texttt{foot}, \texttt{wrist/hand}, and \texttt{elbow}, exhibits structured within-group transfer (brown region), consistent with shared field-of-view, pose constraints, and coil placement. Table~\ref{tab:fov_res_by_anatomy} supports this interpretation: \texttt{ankle}, \texttt{foot}, \texttt{wrist/hand}, and \texttt{elbow} have comparable in-plane FOVs, and these anatomies are typically prescribed with similar geometric coverage. Transfer is also directional: an \texttt{ankle}-trained model remains competitive across the other members of this group, whereas models trained on the smaller categories are typically less competitive on \texttt{ankle}, reflecting the combined role of anatomical similarity and data volume.

Finally, low-data anatomies such as \texttt{pelvis} and \texttt{tib/fib} (pink area) are not best by training on themselves. \texttt{Hip}-trained models outperform \texttt{pelvis}-only training and also transfer strongly to \texttt{tib/fib}, suggesting that these categories are bottlenecked by limited sample size. Notably, \texttt{hip} and \texttt{pelvis} also transfer well to each other despite differences in dataset size, consistent with shared anatomy and acquisition geometry.

\begin{table}[ht]
\begin{minipage}[t]{0.48\textwidth}  

\PA{Table~\ref{tab:mixed_arch} shows the same mixed-anatomy comparison for U-Net and ViT on knee and shoulder reconstruction. Training on the full \methodname{} dataset improves over training only on the target anatomy. This suggests that anatomical diversity provides complementary supervision across architectural families.}

\end{minipage}\hfill
\begin{minipage}[t]{0.5\textwidth}

\centering
\small
\setlength{\tabcolsep}{5pt}
\footnotesize    
\caption{\PA{PSNR (dB) / SSIM for reconstruction with mixed-anatomy training across architectures.}}
\label{tab:mixed_arch}

\resizebox{\linewidth}{!}{
\begin{tabular}{llcc}
\toprule
Model & Test anatomy & Target anatomy only & Full \methodname{} \\
\midrule
U-Net & Knee     & 37.35 / 0.894 & \textbf{37.94 / 0.898} \\
U-Net & Shoulder & 42.05 / 0.952 & \textbf{42.41 / 0.954} \\
ViT   & Knee     & 34.01 / 0.850 & \textbf{37.47 / 0.892} \\
ViT   & Shoulder & 35.79 / 0.912 & \textbf{41.66 / 0.950} \\
\bottomrule
\end{tabular}
}
\end{minipage}%
\end{table}


\subsection{Robustness to Acquisition Protocol}

\begin{figure}[ht]
    \centering
    \includegraphics[width=0.62\linewidth]{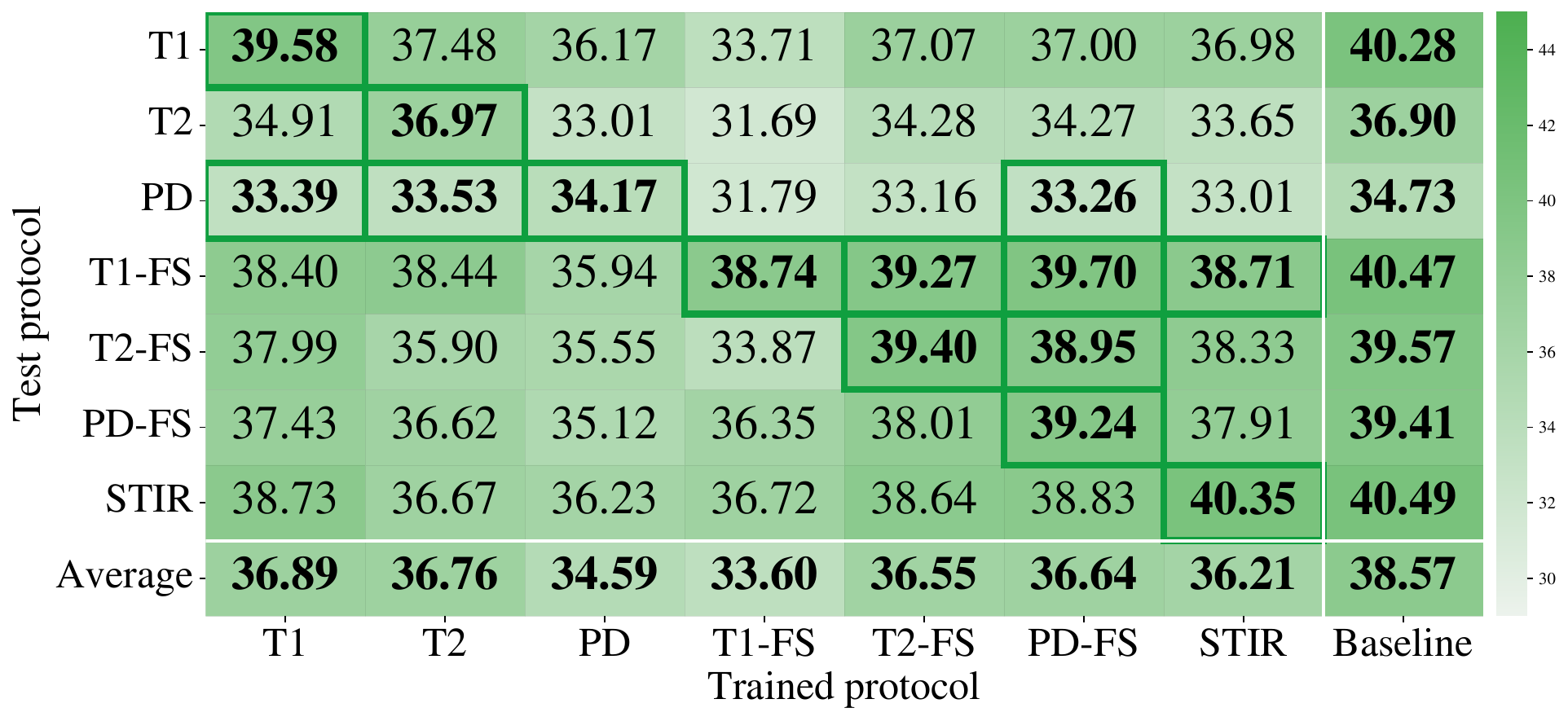}
\caption{Protocol generalization in \methodname{} with E2E-VarNet trained on each protocol (columns) and tested on each protocol (rows); \textbf{Baseline} is trained on all protocols. Boxes mark single-protocol models within 1~dB of the best per row.}
\label{fig:protocol_heatmap}
\end{figure}

Some cross-anatomy transfer patterns may reflect differences in protocol composition. To isolate this, we train VarNet on each contrast group and evaluate across all groups (Fig.~\ref{fig:protocol_heatmap}). Single-protocol training is often brittle under protocol shift: fat-suppressed protocols transfer best to other fat-suppressed targets, while performance typically drops on non--fat-suppressed targets. The model trained on all contrasts provides the strongest whole-test average.

\begin{figure}[ht]
    \centering
    \includegraphics[width=0.70\linewidth]{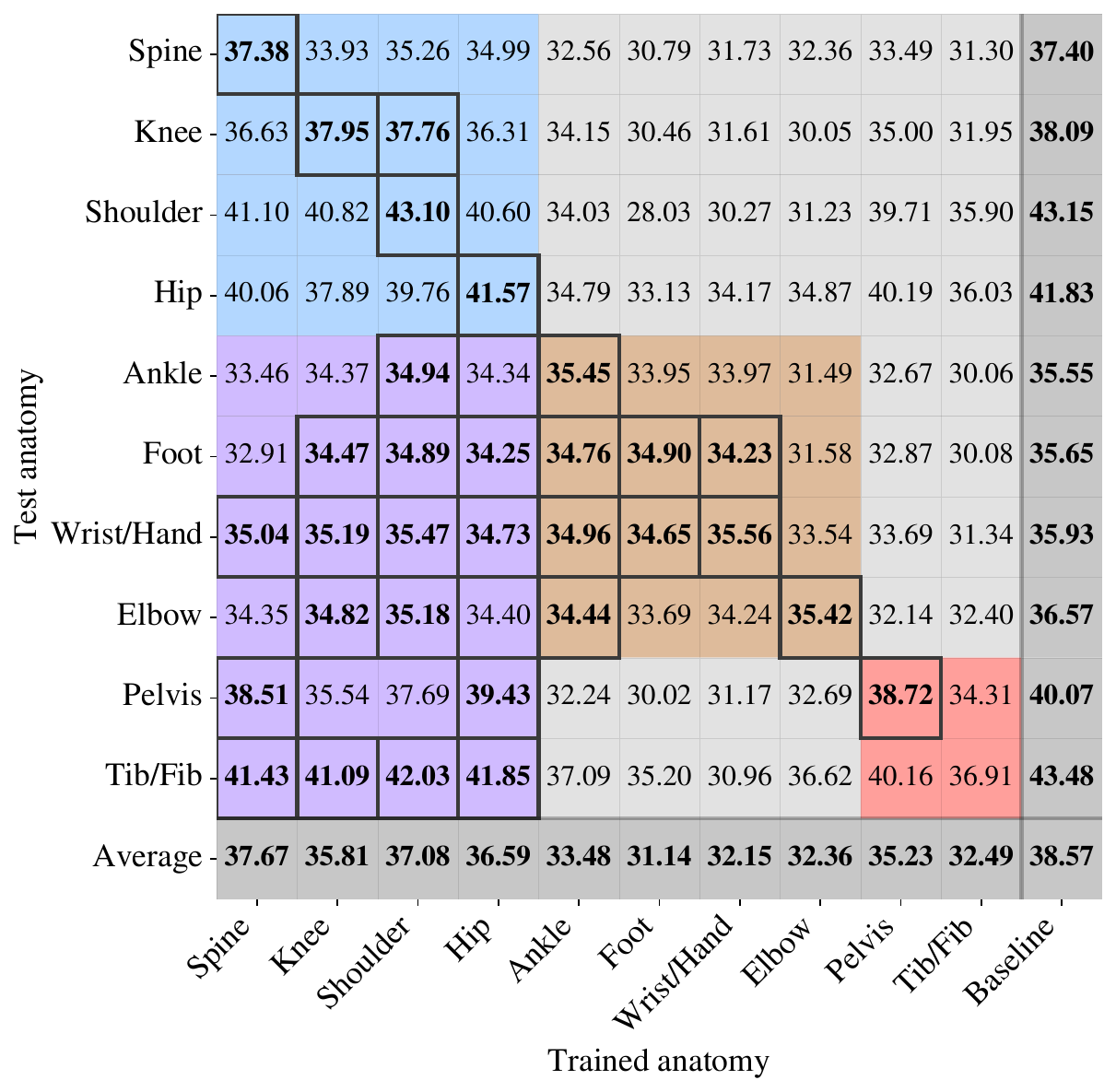}
\caption{\footnotesize Mean PSNR (dB) of E2E-VarNet for cross-anatomy transfer on \textsc{\methodname{}}. Rows are test anatomies, and columns are the training anatomy; the final \textit{Baseline} column corresponds to the model trained on all data. Anatomies are ordered from higher to lower volume counts (top-left to bottom-right), yielding three groups: high-data anchors (blue), distal extremities (brown), and low-data groups (pink). For each test anatomy (row), black outlines mark all anatomy models within 1~dB of the best anatomy result for that row (excluding the baseline).}
\label{fig:anatomy_robustness}
\end{figure}

Protocol distribution also provides context for the anatomy-level transfer patterns in Fig.~\ref{fig:anatomy_robustness}. The appendix (Fig.~\ref{fig:diversity_by_anatomy}) shows that some anatomies (e.g., \texttt{ankle} and \texttt{hip}) are acquired with a relatively diverse mix of contrasts rather than a single dominant protocol. This broader protocol coverage can improve cross-anatomy generalization despite these anatomies having fewer volumes than \texttt{spine} or \texttt{knee}, which generalization performs similarly. Conversely, contrasts that are common across anatomies (e.g., T1, PD-FS, T2-FS) tend to generalize better across protocols, consistent with their broader coverage in the dataset.


%

\subsection{Robustness to Dataset Shift} \label{sec:robustness_dataset_shift}

To test robustness to dataset shift, we train E2E-VarNet (8 cascades, $8\times$ acceleration) on fastMRI, on the full \methodname{} training set, and on \methodname{}-knee, and evaluate each model on the fastMRI validation set, the \methodname{} test set, and the \methodname{} \texttt{knee} test set (Table~\ref{tab:fmri_mmri_knee_cross_dataset}).

\begin{table}[ht]
\begin{minipage}[t]{0.58\textwidth}
\raggedright
The fastMRI-trained model reaches 36.90~dB on fastMRI, while the \methodname{}-trained model achieves 38.57~dB on \methodname{}. Out-of-domain performance drops: fastMRI$\rightarrow$\methodname{} reaches 34.76~dB, whereas \methodname{}$\rightarrow$fastMRI reaches 35.89~dB, 1.01~dB below the fastMRI in-domain result. On the \methodname{} \texttt{knee} test set, the full and knee-only \methodname{} models obtain 38.09 and 37.95~dB, respectively, while the knee-only model reaches 33.13~dB on fastMRI.

These results suggest that training on a diverse MSK dataset improves robustness, while narrower training distributions can fail under shifts in anatomy and acquisition protocol.
\end{minipage}\hfill
\begin{minipage}[t]{0.40\textwidth}
\centering
\setlength{\tabcolsep}{1.8pt}
\renewcommand{\arraystretch}{1.0}
\caption{Cross-dataset performance in PSNR (dB) and SSIM. The \textsc{\methodname{} (knee)} model is trained only on the \texttt{knee} subset of \methodname{}. Values in bold represent the best model by test set.}
\label{tab:fmri_mmri_knee_cross_dataset}
\resizebox{\linewidth}{!}{
\begin{tabular}{lccc}
\toprule
 & \multicolumn{3}{c}{\textbf{Model trained on}} \\
\cmidrule(lr){2-4}
\textbf{Test set} & \textbf{fastMRI} & \textbf{\methodname{}} & \textbf{\shortstack{\methodname{}\\(knee)}} \\
\midrule
\textbf{fastMRI} & \textbf{36.90 / 0.888} & 35.89 / 0.876 & 33.13 / 0.832 \\
\midrule
\textbf{\methodname{}} & 34.76 / 0.886 & \textbf{38.57 / 0.924} & 35.81 / 0.889 \\
\midrule
\textbf{\begin{tabular}[c]{@{}l@{}}\methodname{}\\(knee)\end{tabular}} & 33.04 / 0.849 & \textbf{38.09 / 0.891} & 37.95 / 0.890 \\
\bottomrule
\end{tabular}
}
\end{minipage}%
\end{table}

\subsection{Data Filtering Experiments} \label{sec:data_filtering}

\PA{The previous sections show that mixed-anatomy training improves reconstruction, but they do not isolate why it helps. We therefore test whether the gain is explained only by target-anatomy or target-contrast matching, or whether visually related examples from across the dataset also provide useful supervision. For knee and shoulder PD-FS, we train E2E-VarNet on target-anatomy data, target-contrast data, anatomy-and-contrast data, a random mixed-anatomy subset, a DreamSim-selected subset \citep{fu2023dreamsim}, and the full \methodname{} training split. For DreamSim selection, each target slice retrieves its top-$k$ matches from all eligible training slices; the candidates are merged, deduplicated, ranked by similarity, and truncated to the same slice budget as the corresponding anatomy-and-contrast subset: 541 slices for knee and 1,744 slices for shoulder. All models are evaluated on the corresponding test anatomy and contrast.}

\begin{table}[ht]
\centering
\scriptsize
\setlength{\tabcolsep}{3pt}
\caption{\PA{Controlled-subset PSNR (dB) / SSIM and training-slice counts. The vertical rule separates the broad/full-training comparison from the slice-budget-matched comparison; bold denotes the best result within each three-model group.}}
\label{tab:controlled_subsets}
\resizebox{\linewidth}{!}{
\begin{tabular}{lccc|ccc}
\toprule
& \multicolumn{3}{c|}{Broad/full training}
& \multicolumn{3}{c}{Matched slice budget} \\
\cmidrule(lr){2-4}\cmidrule(lr){5-7}
Test anatomy (contrast)
& Anatomy-only
& Contrast-specific
& \methodname{}
& Slice-matched random
& Anatomy+contrast
& DreamSim \\
\midrule
Knee (PD-FS)
& 34.41 / 0.802
& 34.54 / 0.803
& \textbf{34.61 / 0.806}
& 32.95 / 0.781
& 33.40 / 0.787
& \textbf{33.75 / 0.792} \\
\emph{Training slices}
& 5,864 & 8,661 & 61,903 & 541 & 541 & 541 \\
\cmidrule(lr){1-7}
Shoulder (PD-FS)
& \textbf{44.64 / 0.974}
& 44.39 / 0.973
& 44.63 / \textbf{0.974}
& 42.96 / 0.968
& 42.81 / 0.968
& \textbf{43.63 / 0.971} \\
\emph{Training slices}
& 8,534 & 8,661 & 61,903 & 1,744 & 1,744 & 1,744 \\
\bottomrule
\end{tabular}
}

\end{table}

\PA{Table~\ref{tab:controlled_subsets} shows that anatomy or contrast matching alone does not explain the benefit of broader training data. Among the broad/full-training models, full \methodname{} performs best on knee PD-FS (34.61~dB/0.806), while anatomy-only training is marginally best on shoulder PD-FS (44.64~dB/0.974), exceeding the full model by only 0.01~dB and tying it in rounded SSIM. Under matched training-slice budgets, DreamSim performs best for both anatomies. On knee, it reaches 33.75~dB/0.792, improving over random selection by 0.80~dB/0.011 and anatomy-and-contrast selection by 0.35~dB/0.005. On shoulder, it reaches 43.63~dB/0.971, improving over random selection by 0.67~dB/0.003 and anatomy-and-contrast selection by 0.82~dB/0.003. These results support our hypothesis that visual similarity can identify useful cross-anatomy supervision that is not captured by anatomy and contrast labels alone.}

\subsection{\PA{Zero-shot Transfer to low-field set}} \label{sec:lowfield_zeroshot}

\PA{We test zero-shot transfer on the 0.55~T low-field set (more details in Appendix \ref{apx:lowfield_set}) at $8\times$ acceleration, as shown in Table~\ref{tab:lowfield_zeroshot} and Appendix Fig.~\ref{fig:lowfield_distribution}. Performance drops under field-strength shift, but using more 1.5~T training data generally improves zero-shot performance across architectures. The ViT trained on full \methodname{} obtains the best average performance (30.169~dB/0.782) and the best results on knee and spine.}

\begin{table}[ht]
\centering
\begin{minipage}{\linewidth}
\scriptsize
\setlength{\tabcolsep}{3pt}
\caption{\PA{Zero-shot reconstruction on the 0.55~T low-field set at $8\times$ acceleration. Each cell reports PSNR (dB) / SSIM; bold marks the best metric within each architecture and test-set column.}}
\label{tab:lowfield_zeroshot}

\centering
\resizebox{0.95\linewidth}{!}{%
\begin{tabular}{llcccc}
\toprule
& & \multicolumn{4}{c}{\PA{\textbf{Test set}}} \\
\cmidrule(lr){3-6}
\PA{\textbf{Model}} &\PA{\textbf{Training data}} & \PA{\textbf{Knee}} & \PA{\textbf{Shoulder}} & \PA{\textbf{Spine}} & \PA{\textbf{Average}} \\
\midrule
& \PA{\methodname{} (10\%)} & \PA{29.285 / 0.753} & \PA{28.794 / 0.676} & \PA{28.385 / 0.708} & \PA{28.821 / 0.712} \\
\PA{E2E-VarNet} & \PA{\methodname{} (50\%)} & \PA{29.836 / 0.773} & \PA{29.028 / 0.688} & \PA{28.443 / 0.720} & \PA{29.102 / 0.727} \\
& \PA{\methodname{}} & \PA{\textbf{30.184 / 0.788}} & \PA{\textbf{29.382 / 0.711}} & \PA{\textbf{29.202 / 0.757}} & \PA{\textbf{29.590 / 0.752}} \\
\midrule
& \PA{\methodname{} (10\%)} & \PA{29.759 / \textbf{0.805}} & \PA{28.449 / 0.712} & \PA{29.205 / 0.777} & \PA{29.093 / 0.762} \\
\PA{U-Net} & \PA{\methodname{} (50\%)} & \PA{29.855 / 0.798} & \PA{28.858 / 0.715} & \PA{29.564 / 0.780} & \PA{29.396 / 0.762} \\
& \PA{\methodname{}} & \PA{\textbf{30.050} / 0.804} & \PA{\textbf{28.902 / 0.717}} & \PA{\textbf{29.870 / 0.786}} & \PA{\textbf{29.579 / 0.767}} \\
\midrule
& \PA{\methodname{} (10\%)} & \PA{28.668 / 0.777} & \PA{27.940 / 0.687} & \PA{27.130 / 0.715} & \PA{27.842 / 0.722} \\
\PA{ViT} & \PA{\methodname{} (50\%)} & \PA{30.549 / 0.813} & \PA{\textbf{30.106} / 0.736} & \PA{28.990 / 0.756} & \PA{29.816 / 0.765} \\
& \PA{\methodname{}} & \PA{\textbf{30.724 / 0.825}} & \PA{29.952 / \textbf{0.744}} & \PA{\textbf{29.958 / 0.786}} & \PA{\textbf{30.169 / 0.782}} \\
\bottomrule
\end{tabular}}
\end{minipage}
\end{table}

\section{Discussion} \label{sec:discussion}

We introduce \methodname{}, a raw MSK MRI dataset and benchmark with 3,096 fully sampled 1.5~T multi-coil volumes (88,416 slices) and a 54-volume 0.55~T evaluation set (1,414 slices). Using accelerated reconstruction as a testbed, we benchmark classical, unrolled, convolutional, transformer, and generative baselines, and study robustness under anatomy, protocol, dataset, and field-strength shifts.

Across E2E-VarNet, U-Net, and ViT baselines, training on anatomically and protocol-diverse MSK data consistently improves reconstruction performance compared to anatomy-specific training. Our experiments show that reconstruction robustness and transferability depend jointly on anatomy, acquisition protocol, training-set scale, and acquisition geometry. Cross-anatomy transfer follows structured patterns, with related anatomies providing useful supervision in limited-data settings. Zero-shot evaluation on the 0.55~T low-field set shows a substantial field-strength shift, with the full-data E2E-VarNet obtaining $29.66\pm3.06$~dB and $0.759\pm0.082$ SSIM. Similarity-aware filtering experiments indicate that compact cross-anatomy subsets can retain useful cross-anatomy supervision under matched training budgets, suggesting a promising direction for efficient dataset design.

\section{Limitations} \label{sec:limitations}

Although \methodname{} is the largest publicly available raw MSK MRI dataset to date, the current release has several limitations. The 1.5~T core dataset is collected from a single site and vendor, so our results should not be interpreted as demonstrating full robustness to multi-site or multi-vendor shifts. The 0.55~T low-field set is also relatively small, and is used only for zero-shot evaluation rather than supervised low-field training.
Our experiments focus on accelerated reconstruction with retrospective undersampling and standard quantitative metrics. While these experiments provide a controlled benchmark for robustness and transfer, they do not directly measure downstream diagnostic performance or clinical reader preferences. Future work should expand the dataset across sites, vendors, and field strengths, increase the scale and diversity of low-field MSK acquisitions, and evaluate reconstruction failures, clinical tasks, and reader studies before deployment in prospective clinical settings.

\section*{Acknowledgements}
This work was primarily supported by NIH Award DP2LM014564-01. This work was partially supported by AWS credits through an Amazon Faculty Research Award, a NAIRR Pilot Award, and generous funding by Coefficient Giving. M. Soltanolkotabi is supported by the Packard Fellowship in Science and Engineering, a Sloan Research Fellowship in Mathematics, NSF CAREER Award \#1846369, DARPA FastNICS program, NSF CIF Awards \#1813877 and \#2008443.



\bibliographystyle{ieeenat_fullname}
\bibliography{references, references2}

\newpage
\appendix
\section{Appendix}

\subsection{{0.55~T Low-Field Set}} \label{apx:lowfield_set}
{The low-field set contains 54 scans (1,414 slices) from 8 patients across knee, shoulder, and spine.}
\begin{figure}[ht]
    \centering
    \includegraphics[width=0.92\linewidth]{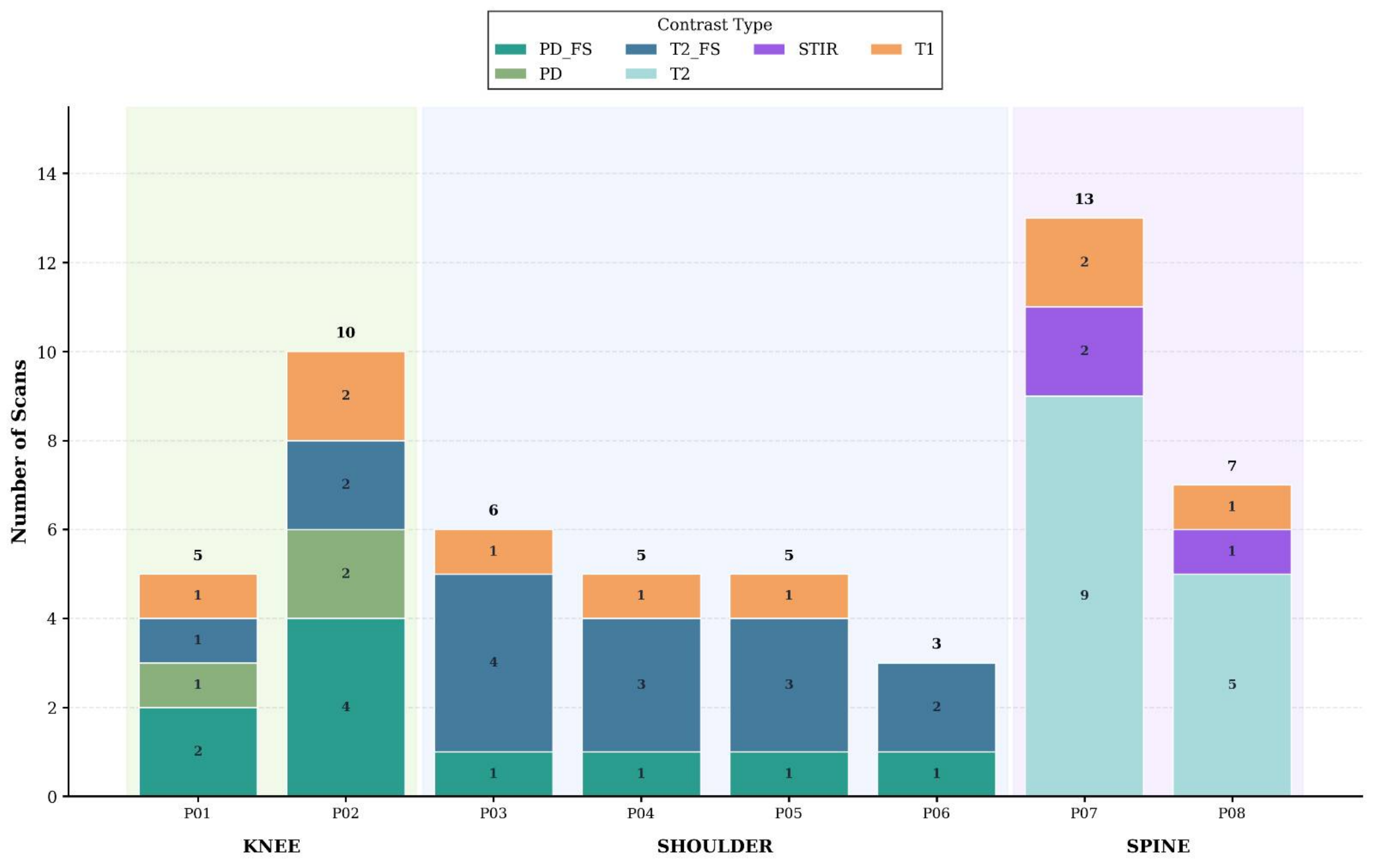}
    \caption{{Distribution of the 0.55~T low-field set by patient, anatomy, and contrast.}}
    \label{fig:lowfield_distribution}
\end{figure}

\subsection{Accelerated MRI fundamentals} \label{apx:acc_mri_primer}
In Magnetic Resonance Imaging (MRI), anatomical information is acquired in the Fourier domain (commonly {referred} to as \textit{k-space}) using receiver coils. In parallel imaging, an array of $N$ receiver coils is employed, where each coil observes spatially different parts of the underlying image $\bm{x}^* \in \mathbb{C}^n$ modulated by its own complex-valued spatial sensitivity map $S_i$. The measurement obtained by the $i$-th coil is modeled as:
\begin{equation*}
    \bm{k}_i = \mathcal{F} S_i \bm{x}^* + \bm{z}_i, \quad i = 1, \dots, N
\end{equation*}
where $\mathcal{F}$ denotes the two-dimensional Fourier transform and $\bm{z}_i$ represents additive measurement noise. Acquiring fully-sampled k-space data is time consuming. To reduce scan time, accelerated MRI techniques undersample k-space. This undersampling process is mathematically described by applying a binary sampling mask $\bm{M}$ to the full sampled k-space data:
\begin{equation*}
    \tilde{\bm{k}}_i = \bm{M} \bm{k}_i, \quad i = 1, \dots, N
\end{equation*}
The mask $\bm{M}$ sets unacquired frequency locations to zero, thereby reducing the amount of data collected. By stacking the measurements from all coils, the forward model can be compactly expressed as
\begin{align*}
    \tilde{\bm{k}} = \mathcal{A}\rbr{\bm{x}^*}
\end{align*} 
where $\mathcal{A}\rbr{\cdot}$ denotes the linear forward operator and $\tilde{\bm{k}}$ represents the aggregated undersampled k-space measurements. The objective of accelerated MRI reconstruction is to recover the unknown image $\bm{x}^*$ from $\tilde{\bm{k}}$. 

However, due to undersampling below the Nyquist rate, perfect recovery is generally impossible without additional assumptions. Problems of this form fall under the framework of compressed sensing (CS). Classical CS approaches for accelerated MRI rely on prior structural assumptions about the underlying image $\bm{x}^*$, such as sparsity in a transform domain. Under this framework, image recovery is formulated as a convex optimization problem:
\begin{align} \label{eq:opt}
    \hat{\bm{x}} = \argmin_{\bm{x}} \twonorm{\mathcal{A}\rbr{\bm{x}} - \tilde{\bm{k}}}^2 + \mathcal{R}\rbr{\bm{x}}
\end{align}
where $\mathcal{R}\rbr{\cdot}$ denotes a regularization term that promotes sparsity in a chosen domain. Common regularizers used in CS based MRI reconstruction include $\ell_1$-norm penalties in the wavelet domain and total-variation (TV). These problems are typically solved numerically using iterative methods based on gradient descent.

\subsection{Additional experimental details} \label{apx:exp_details}

\textbf{ESPIRiT} \citep{uecker_espirit_2014} -- We run ESPIRiT through \texttt{run\_bart.py} script available through GitHub repository of fastMRI\footnote{https://github.com/facebookresearch/fastMRI}. We run the script for 200 iterations per slice with TV regularization strength set to $10^{-2}$.

\textbf{E2E-VarNet} \citep{sriram_end--end_2020} -- Unless specified, we use the default hyperparameters and the model architecture available through fastMRI's GitHub codebase. For the scaling law experiments in Section \ref{sec:scaling_laws}, we change the number of cascades in the model from $\bbr{4,8,12}$.

\textbf{LBM} \citep{chadebec2025lbmlatentbridgematching} -- We set the source and target image distributions to be zero-filled reconstruction of the undersampled measurement and RSS reconstruction respectively.

Experiments were run on NVIDIA A100 80GB GPUs (Amazon EC2) and NVIDIA H100 80GB GPUs, using a per-GPU batch size of 1.

Training VarNet with 8 cascades on the full \methodname{} training split required approximately 6.5 hours on a single H100 GPU. The longest configuration---16 cascades trained on the full dataset---took approximately 10.5 hours on a single H100 GPU.

\subsection{Additional \methodname{} statistics} \label{apx:dataset_statistics}

The public release excludes demographic and identifying fields, including age, sex, height, weight, birth date, patient name, original identifiers, and study date. We therefore do not report demographic statistics derived from private source metadata.

{\raggedright All paper analyses use the ten public anatomy classes separately: spine, shoulder, hip, knee, wrist/hand/thumb, ankle, foot, pelvis/pubalgia, elbow, and leg/tib-fib. Coarse and finer labels remain available as complementary metadata but are not substituted for the ten-class benchmark labels.\par}

\begin{figure}[ht]
    \centering
    \includegraphics[width=1.0\linewidth]{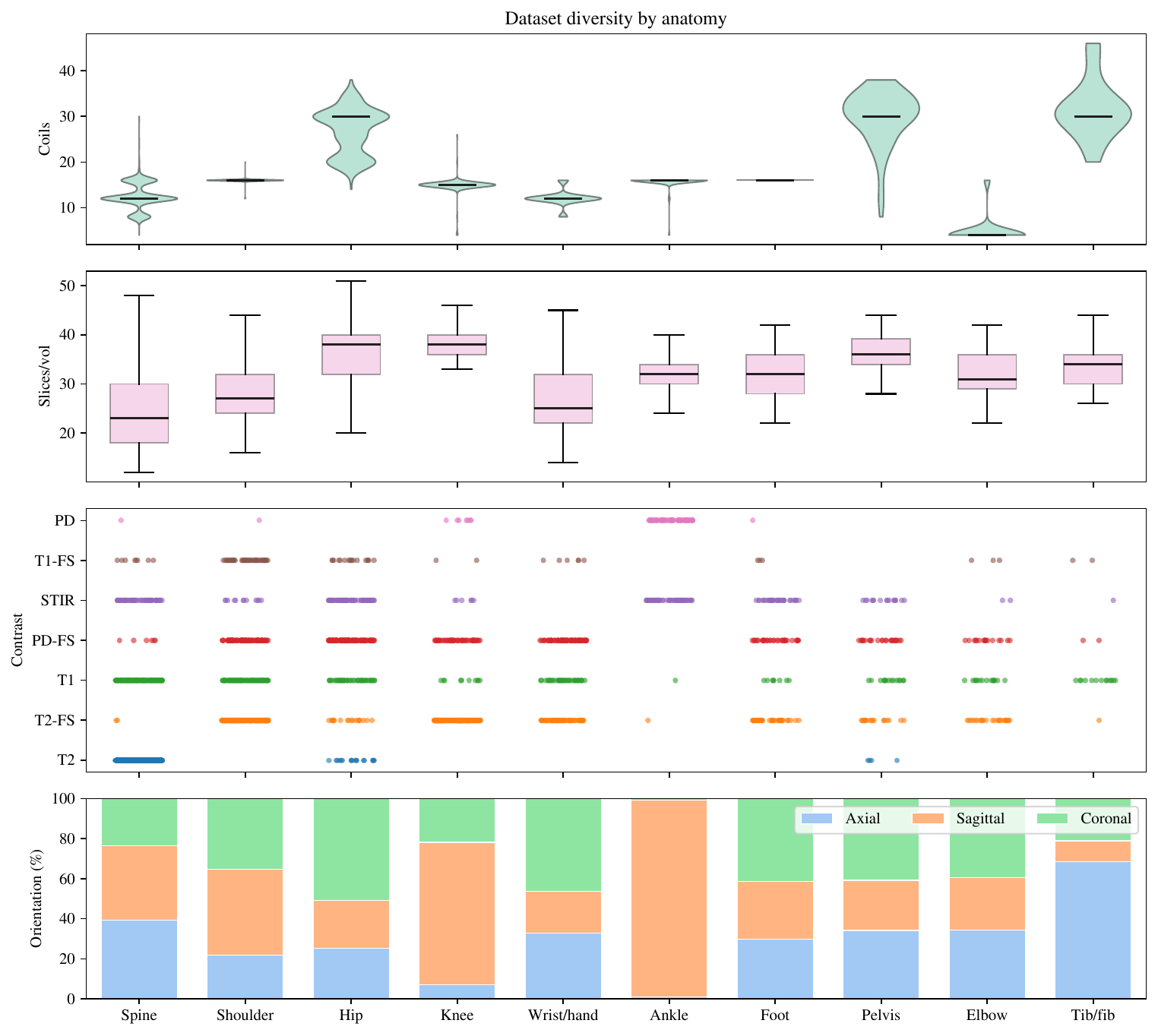}
    \caption{Dataset diversity by anatomy. Violin/box plots summarize the distribution of receive-coil counts and slices per volume for each anatomy group. A contrast panel reports the presence of major contrast families (T1, T1-FS, T2, T2-FS, PD, PD-FS, STIR) per anatomy, reflecting protocol heterogeneity. A stacked bar chart shows the orientation mix (axial/sagittal/coronal) within each anatomy, highlighting anatomy-dependent acquisition geometry.}
    \label{fig:diversity_by_anatomy}
\end{figure}

\begin{table}[!ht]
\centering
\small
\setlength{\tabcolsep}{5pt}
\renewcommand{\arraystretch}{1.15}
\caption{Average in-plane field-of-view (FOV) and in-plane resolution by anatomy.}
\label{tab:fov_res_by_anatomy}
\resizebox{\linewidth}{!}{
\begin{tabular}{lcccccccccc}
\toprule
\textbf{Metric} & \textbf{spine} & \textbf{knee} & \textbf{shoulder} & \textbf{hip} & \textbf{ankle} & \textbf{foot} & \textbf{wrist/hand} & \textbf{elbow} & \textbf{pelvis} & \textbf{tib/fib} \\
\midrule
\textbf{FOV$_x$ (mm)} & 238.28 & 159.42 & 162.69 & 281.82 & 165.86 & 155.96 & 109.46 & 167.38 & 305.26 & 257.37 \\
\textbf{FOV$_y$ (mm)} & 237.46 & 156.31 & 162.43 & 278.46 & 166.59 & 150.62 & 99.13 & 166.36 & 294.00 & 243.40 \\
\textbf{Res$_x$ (mm)} & 0.7016 & 0.3674 & 0.3076 & 0.7881 & 0.4483 & 0.4364 & 0.3185 & 0.5200 & 0.8628 & 0.7128 \\
\textbf{Res$_y$ (mm)} & 0.7073 & 0.4138 & 0.4254 & 0.8983 & 0.4483 & 0.4527 & 0.3319 & 0.5620 & 0.9485 & 0.8700 \\
\bottomrule
\end{tabular}
}
\end{table}

Figure~\ref{fig:diversity_by_anatomy} summarizes key sources of acquisition heterogeneity across these anatomy groups by reporting: (i) the distribution of receive-coil counts; (ii) the distribution of slices per volume; (iii) the observed contrast families (T1/T2/PD with and without fat suppression, plus STIR) across anatomies, highlighting protocol mix; and (iv) the per-anatomy distribution of scan orientation (axial/sagittal/coronal). In addition, Table~\ref{tab:fov_res_by_anatomy} reports average in-plane field-of-view (FOV) and in-plane resolution by anatomy, showing that acquisition geometry also varies substantially in physical scale (FOV$_x$/FOV$_y$) and sampling density (Res$_x$/Res$_y$). 

\subsection{U-Net robustness across anatomy and contrast}
We repeat the anatomy-only and contrast-only whole-test evaluations with U-Net. As with E2E-VarNet, narrow subsets transfer unevenly, while full-data training provides the strongest aggregate result.
\begin{figure}[ht]
    \centering
    \begin{minipage}{0.56\linewidth}
        \includegraphics[width=\linewidth]{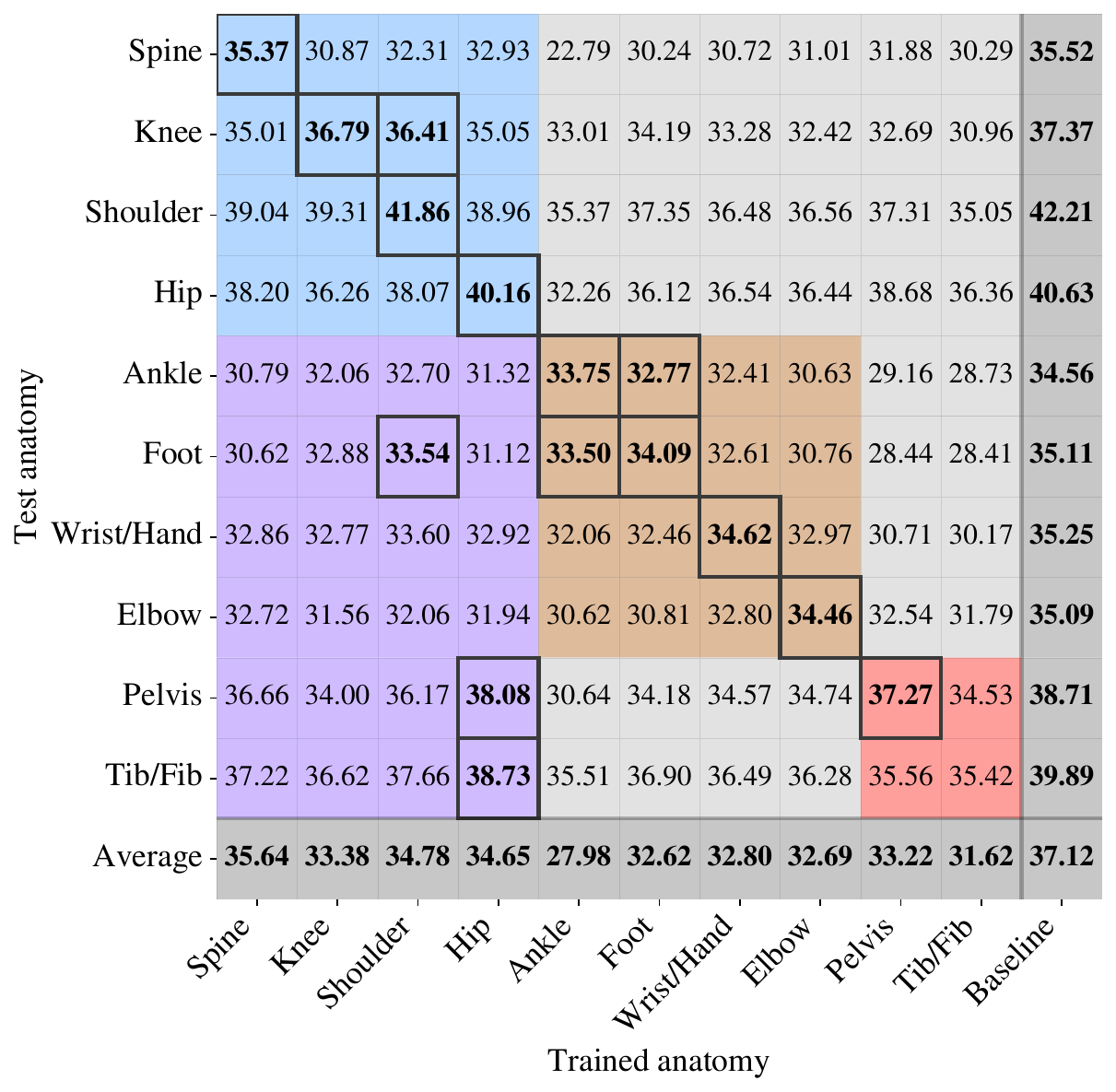}
    \end{minipage}\hfill
    \begin{minipage}{0.42\linewidth}
        \includegraphics[width=\linewidth]{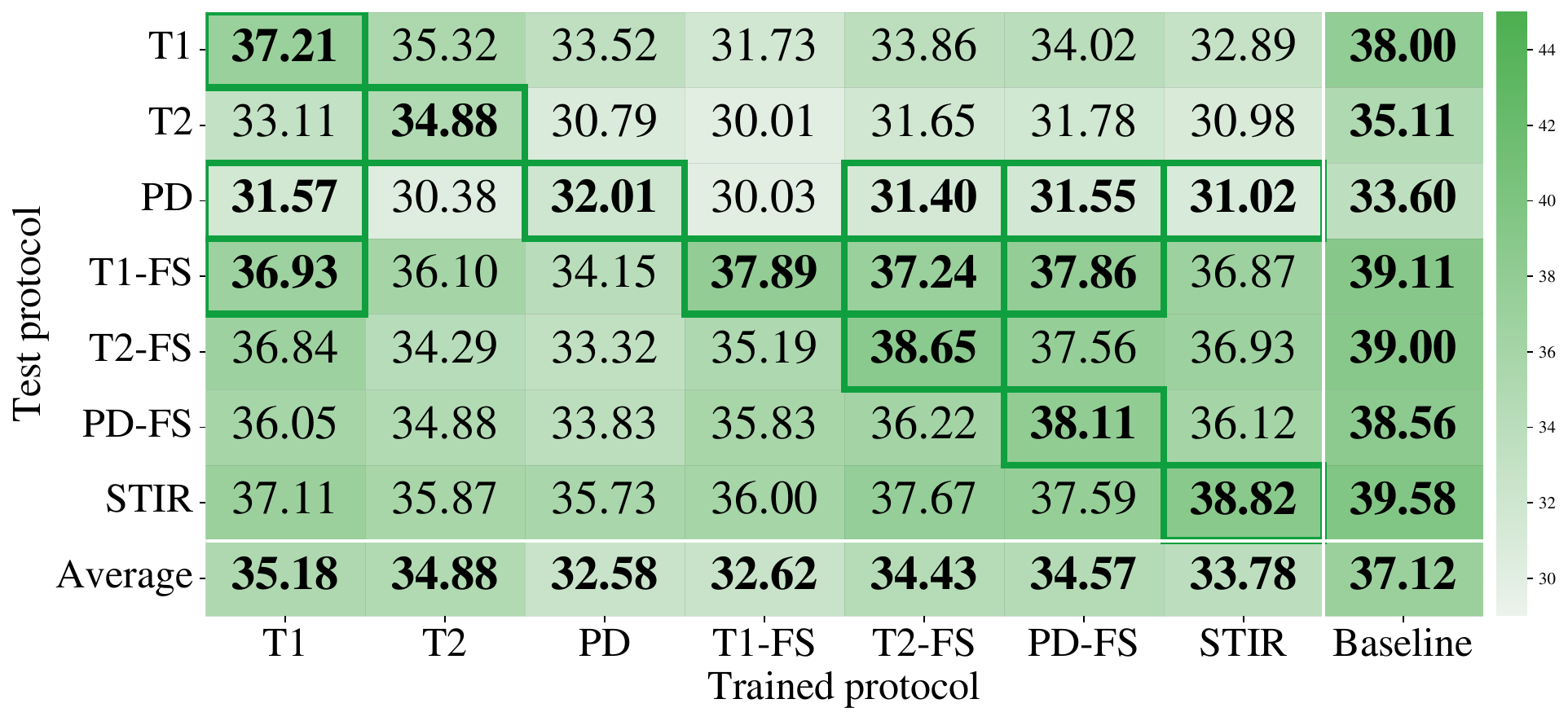}
    \end{minipage}
    \caption{U-Net cross-anatomy (left) and cross-contrast (right) PSNR. Columns are training subsets and rows are test subsets; \textit{All data} is trained on the complete training split.}
    \label{fig:unet_robustness}
\end{figure}

\subsection{Example reconstructions} \label{apx:vis_recon}
\begin{figure}[ht!]
    \centering
    \begin{minipage}[t]{0.49\linewidth}
        \centering
        \includegraphics[width=\linewidth]{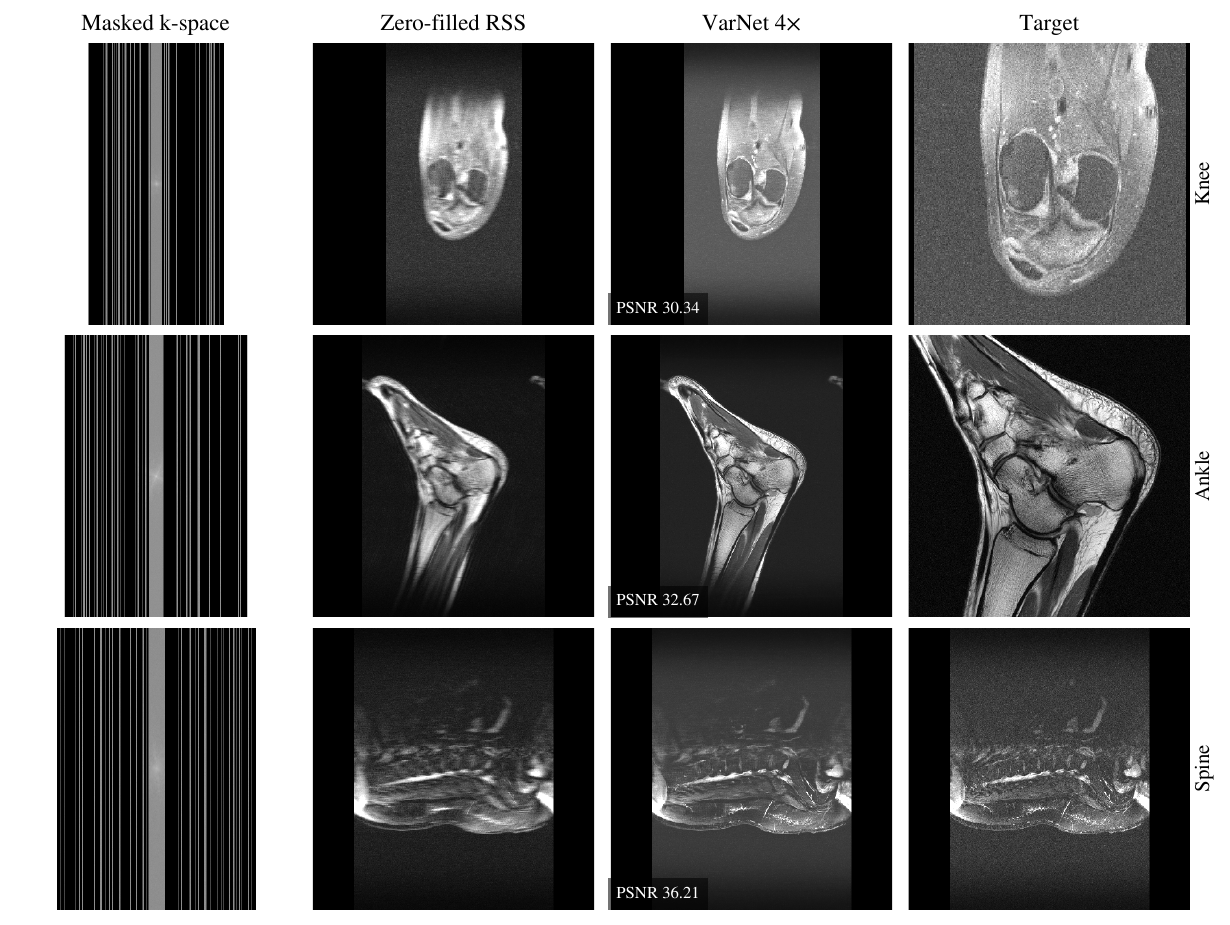}
        \subcaption{\textbf{4$\times$.} Representative slices across anatomies (\texttt{knee}, \texttt{ankle}, \texttt{spine}).}
        \label{fig:recon_visualization_4x}
    \end{minipage}\hfill
    \begin{minipage}[t]{0.49\linewidth}
        \centering
        \includegraphics[width=\linewidth]{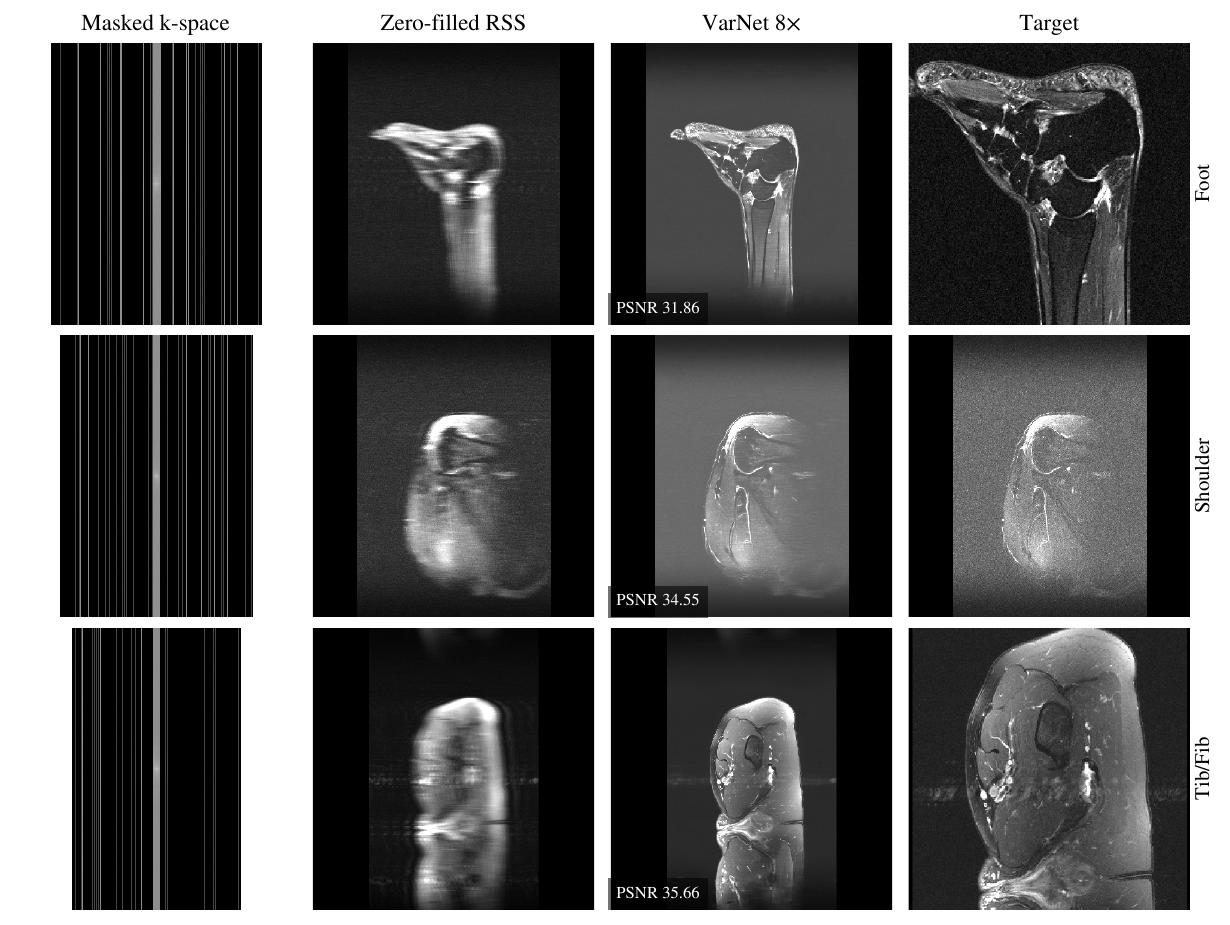}
        \subcaption{\textbf{8$\times$.} Representative slices across anatomies (\texttt{foot}, \texttt{shoulder}, \texttt{tib/fib}).}
        \label{fig:recon_visualization_8x}
    \end{minipage}
    \caption{\textbf{Qualitative accelerated reconstruction examples across anatomies.}
    For each panel, columns (left to right) show the masked k-space after applying the undersampling pattern, the zero-filled RSS reconstruction, the reconstruction produced by VarNet trained on full \methodname{}, and the fully sampled target.}
    \label{fig:recon_visualization_4x_8x}
\end{figure}


\end{document}